# Contrast Limited Adaptive Histogram Equalization (CLAHE) Approach for Enhancement of the Microstructures of Friction Stir Welded Joints

Akshansh Mishra[1]
[1]Centre for Artificial Intelligent Manufacturing Systems, Stir Research Technologies, India
[1]Orcid ID: https://orcid.org/0000-0003-4939-359X

**Abstract:**

Image processing algorithms are finding various applications in manufacturing and materials industries such as identification of cracks in the fabricated samples, calculating the geometrical properties of the given microstructure, presence of surface defects, etc. The present work deals with the application of Contrast Limited Adaptive Histogram Equalization (CLAHE) algorithm for improving the quality of the microstructure images of the Friction Stir Welded joints. The obtained results showed that the obtained value of quantitative metric features such as Entropy value and RMS Contrast value were high which resulted in enhanced microstructure images.

## 1. Introduction

In automated manufacturing industries, image processing systems built around industrial cameras are playing an important role [1-2]. Various image processing algorithms have been used by various researchers for the identification of surface defects in fabricated components and also for studying the variation of mechanical and microstructure properties of the manufactured materials [3-6]. Aryanfar et al. [7] with the help of the percolation-based image processing method quantified the flowability of concrete for the plastic viscosity of cementitious mortar with super absorbent polymer. Peng et al. [8] used the digital image processing technique mesoscale fracture analysis of recycled aggregate concrete. Schottel et al. [9] determined the fiber orientation in Discontinuous fiber reinforced polymers (DicoFRP) like Sheet Molding Compounds (SMC )using state-of-the-art image processing algorithms. Almansoori et al. [10] used a novel image processing approach for fault detection in the aircraft body. The proposed model achieved an accuracy of 98.28 %.

The present research work proposes the implementation of Contrast Limited Adaptive Histogram Equalization (CLAHE) image processing algorithm for enhancing the microstructure images of the weld joints obtained by Friction Stir Welding process.

## 2. Concept of Digital Image

Any image can be represented in terms of a two-dimensional function f(x,y) where x and y are spatial (plane) coordinates and at any pair of coordinates (x,y), the amplitude of f is called the intensity or gray level of the given image at that point. The given image is called a digital image when the intensity values of f and the coordinates (x,y) are all finite and discrete quantities.

The digital image constitutes a finite number of elements called pixels each of which has a particular location and a value as shown in Figure 1.

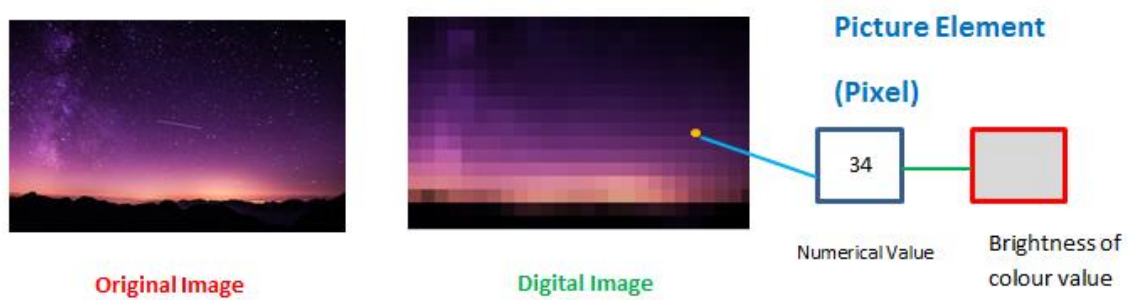

Figure 1: Representation of digital image and its pixel value

Images and other signals are often described by using mathematical models. A signal is often considered as a function depending on some variables and it can be represented in one-dimensional form (i.e. dependent on time), two-dimensional (i.e. an image which depends on the two coordinates of a plane), and three-dimensional (i.e. for tracing the volumetric object in space). Monochromatic image can be described with the help of a scalar function while the color images which consist of three component colors can be described by a vector function.

Functions that are used in image processing applications can be classified as digital, discrete, or continuous. In a continuous function, there is a continuous domain and range while in a discrete function there is a domain set that is discrete in nature, and in a digital function, the range set is discrete in nature. In image processing operation, the nature of the image is considered static in nature i.e. time is constant. Digital image functions are normally used in computerized image processing which is usually represented in the form of matrices, so coordinates are natural numbers. The domain of the image function is a region R in the plane is given by

$$R = \{(x, y), 1 \leq x \leq x_m, 1 \leq y \leq y_n\} \qquad (1)$$

Where $x_m$ and $y_n$ indicates maximal coordinates.

## 3. Concept of Color Space

Color spaces are the organization of the colors in a given image in a specific format. The way in which color is being represented is called a Color model. For effective image representation, various types of color spaces are used by images such as RGB (red, green, blue), XYZ (color in x, y, and z dimensions), HSV/HSL (hue, saturation and value/ hue, saturation and lightness), LCH (Lightness, chroma, and hue), LAB (luminance, and green-red and blue-yellow color components), YPbPr (green, blue and red cables), YUV (brightness and chroma or color), and YIQ (luminance, in-phase parameter, and quadrature).

One color space can be converted into another, for example below is the Python code for converting RGB color space of a brass microstructure to HSV color space.

```
#Reading the microstructure image
from skimage import io
image=io.imread('brass microstructure.jpg')
io.imshow(image)
```

The obtained image is shown in Figure 2.

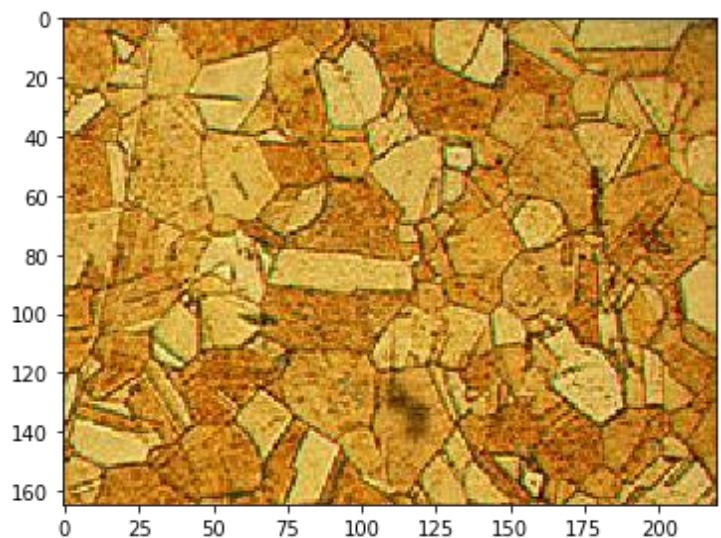

Figure 2: Reading of the image file

```
#Converting RGB color space to HSV color Space and vice versa
from skimage import io
from skimage import color
from skimage import data
from pylab import *
#Read image
img = io.imread('brass microstructure.jpg')
#Convert to HSV
img_hsv = color.rgb2hsv(img)
#Convert back to RGB
img_rgb = color.hsv2rgb(img_hsv)
#Show both figures
figure(0)
io.imshow(img_hsv)
figure(1)
io.imshow(img_rgb)
```

The obtained output image is shown in Figure 3.

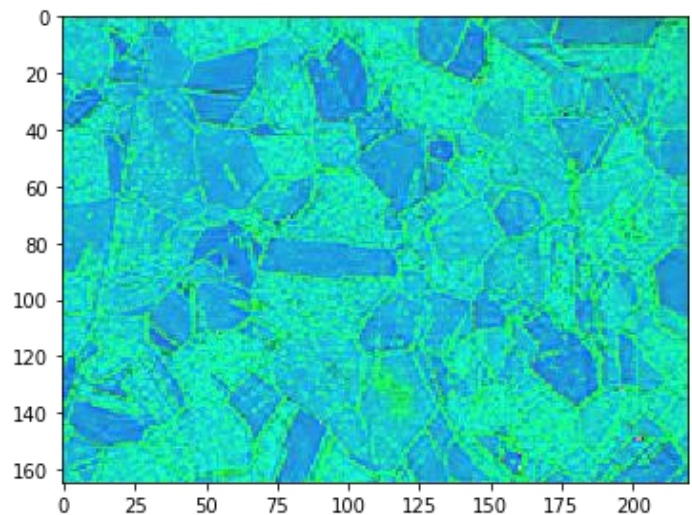

Figure 3: Obtained HSV color space

Connversion of RGB color space to XYZ color space can be done with the help of following Python code.

```
from skimage import io

from skimage import color

from skimage import data

#Read image

img = io.imread('brass microstructure.jpg')

#Convert to XYZ

img_xyz = color.rgb2xyz(img)

#Convert back to RGB

img_rgb = color.xyz2rgb(img_xyz)

#Show both figures

figure(0)

io.imshow(img_xyz)

figure(1)

io.imshow(img_rgb)
```

The obtained output image is shown in Figure 4.

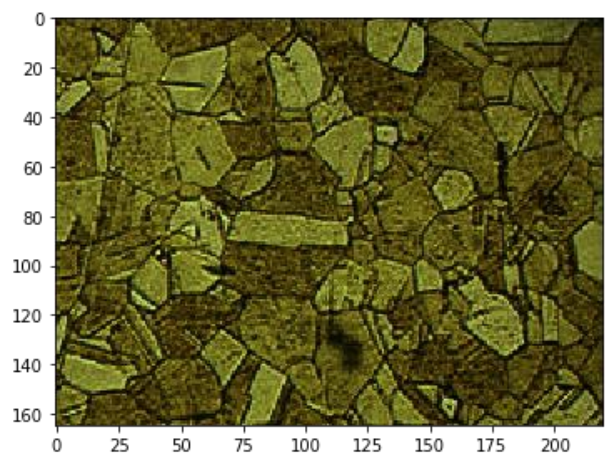

Figure 4: Obtained XYZ color space of the given microstructure

Conversion of RGB color space to LAB color space can be done by following Python code.

```
from skimage import io
from skimage import color
from skimage import data
#Read image
img = io.imread('brass microstructure.jpg')
#Convert to XYZ
img_xyz = color.rgb2xyz(img)
#Convert back to RGB
img_rgb = color.xyz2rgb(img_xyz)
#Show both figures
figure(0)
io.imshow(img_xyz)
figure(1)
io.imshow(img_rgb)
```

The obtained output is shown in Figure 5.

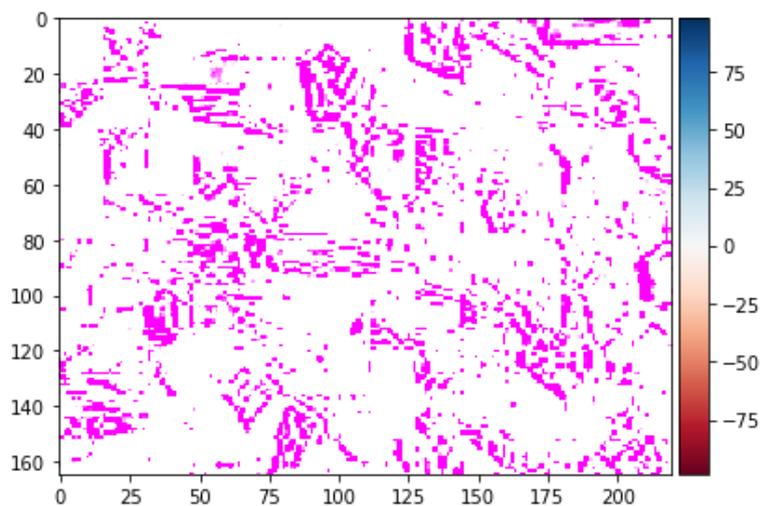

Figure 5: Obtained LAB color space of the given microstructure

Conversion of RGB color space to YUV color space is done with the help of following Python code.

```
from skimage import io
from skimage import color
#Read image
img = io.imread('brass microstructure.jpg')
#Convert to YUV
img_yuv = color.rgb2yuv(img)
#Convert back to RGB
img_rgb = color.yuv2rgb(img_yuv)
#Show both figures
figure(0)
io.imshow(img_yuv)
figure(1)
io.imshow(img_rgb)
```

The obtained output is shown in Figure 6.

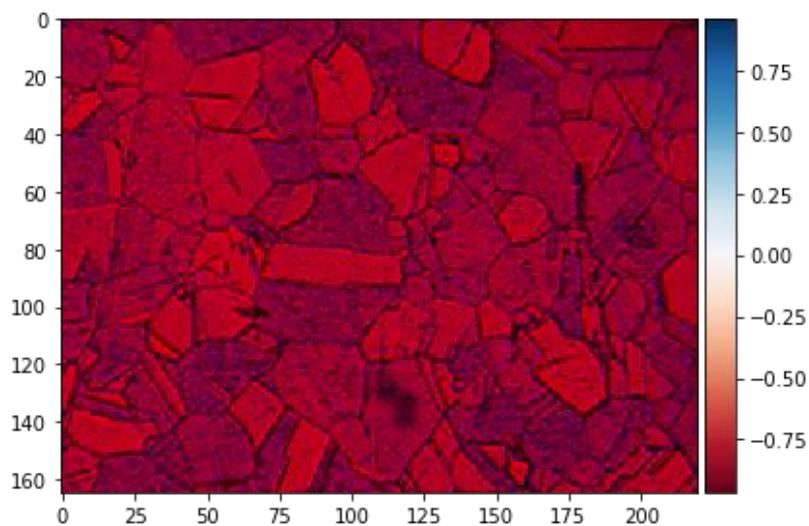

Figure 6: Obtained YUV color space of the given microstructure

Conversion of RGB color space to YIQ color space is done with the help of following Python code.

```
from skimage import io
from skimage import color
#Read image
img = io.imread('brass microstructure.jpg')
#Convert to YIQ
img_yiq = color.rgb2yiq(img)
#Convert back to RGB
img_rgb = color.yiq2rgb(img_yiq)
#Show both figures
figure(0)
io.imshow(img_yiq)
figure(1)
io.imshow(img_rgb)
```

The obtained output image is shown in Figure 7.

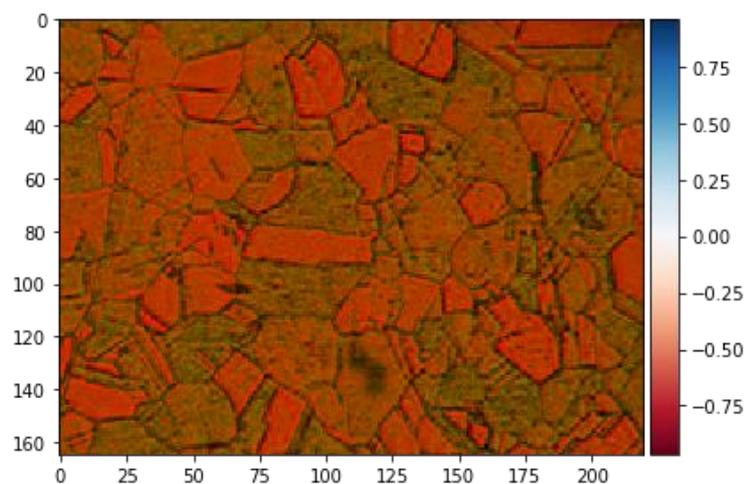

Figure 7: Obtained YIQ color space of the given microstructure

## 4. Experimental Procedure

At the lowest level of abstraction, various operations on images are subjected in which both input and output are intensity images and this operation is called Image pre-processing. The intensity image is commonly represented by a matrix or matrices of the brightness values. The first step is to upload the image on the Google Colaboratory platform and further read the microstructure image. The second step is to subject the microstructure image to a brightness transformation. Generally, brightness transformation depends on the properties of the pixel itself and brightness transformation can be further used to modify the pixel brightness. The pixel brightness transformation can be classified into two types i.e. grayscale transformations and brightness correction. Brightness correction modifies the pixel brightness with regard to its position in the image while grayscale transformation does not modify the pixel brightness with regard to its position in the image. In our present work, we have used grayscale transformation.

The microstructure images used in the present study are shown in Figure 8 and Figure 9.

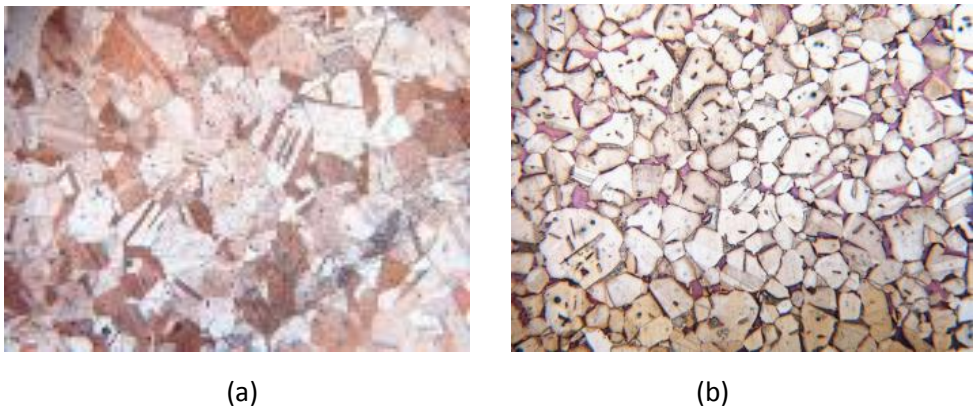

(a)　　　　　　　　　　　　　　(b)

Figure 8: Microstructure images of a) Single phase brass microstructure b) Dual phase brass microstructure[13]

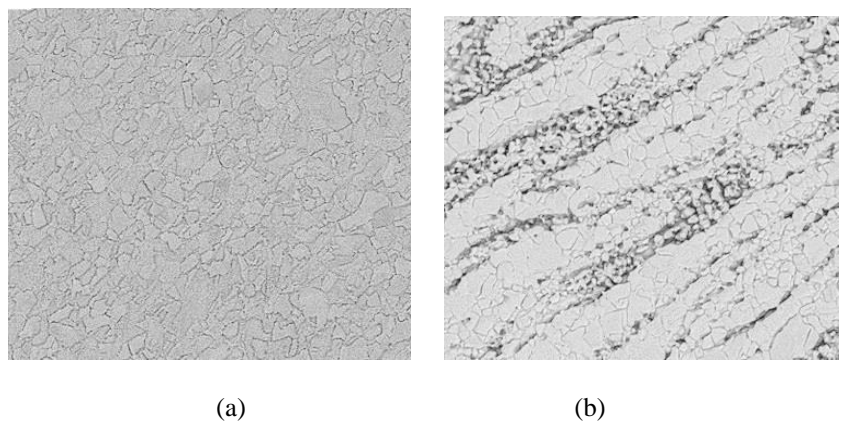

(a)　　　　　　　　　　　　　　(b)

Figure 9: Microstructure image of a) Single phase brass FSWed joint b) Dual phase brass FSWed joint [13]

## 5. Results and Discussion

For assessing the enhancement algorithms, a series of quantitative metrics such as colorfulness metric, entropy, zero channel mean and RMS contrast were used.

The concept of a new color appearance attribute was proposed by RWG Hunt in 1977 [11]. Colorfulness is also referred to as chromaticness [12] which is the attribute of a visual sensation according to which the perceived color of an area appears to be more or less chromatic. The measurement of the intensity of the hue is called the colorfulness of the single color stimulus. The computation of the colorfulness of a given images can be done with the help of equation 2:

$$M = \sqrt{\sigma^2_{rg} + \sigma^2_{yb}} + 0.3 \sqrt{\mu^2_{rg} + \mu^2_{yb}} \qquad (2)$$

Where $\mu$ and $\sigma$ are the mean value and standard deviation of opposite components i.e. 'rg' and 'yb' of the image pixels. The value of opposite components i.e. 'rg' and 'yb' are calculated by using equation 3 and 4.

$$rg = R - G \qquad (3)$$

$$yb = 0.5(R + G) - B \qquad (4)$$

It can be inferred from Table 1 that dual phase brass microstructure has more colorfulness metric in comparison to other microstructures. So it can be concluded that the dual phase brass microstructure has more colorfulness than rest of microstructures.

Entropy is defined as the method of statistical measures that can be used to characterize the texture of the given input image. A particular value is obtained from the entropy function as per equation 5 which represents a certain level of complexity in the corresponding section of an image.

$$Entropy = -\frac{1}{4}\sum_{k=0}^{3}\sum_{i=0}^{L-1}\sum_{j=0}^{L-1} P(i,j;d,\theta_k)log_{10}P(i,j;d,\theta_k) \qquad (5)$$

Table 1 shows the obtained entropy value of the used microstructures in the present research work. It is observed that the dual-phase brass microstructure has a higher entropy value in comparison to the other microstructures. Figure 10 to Figure 13 shows the obtained entropy images of the individual microstructure.

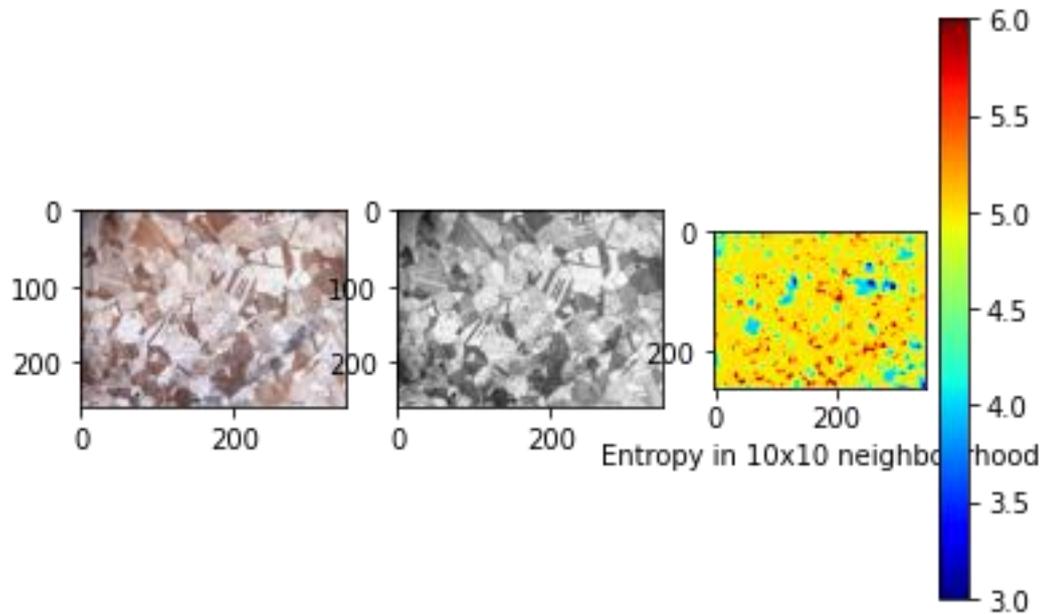

Figure 10: Entropy image of the single-phase brass microstructure

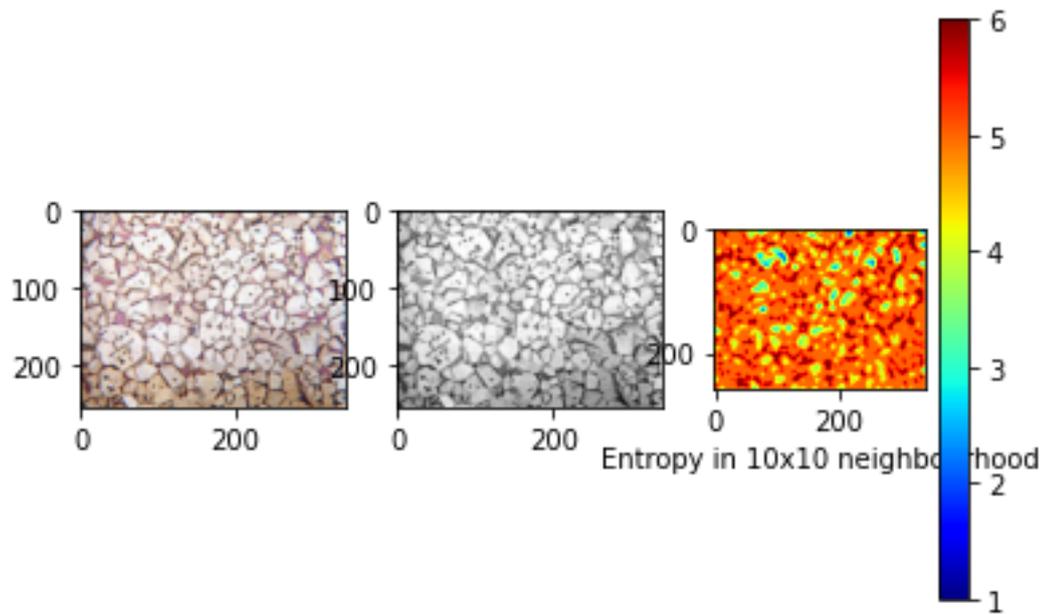

Figure 11: Entropy image of dual-phase brass microstructure

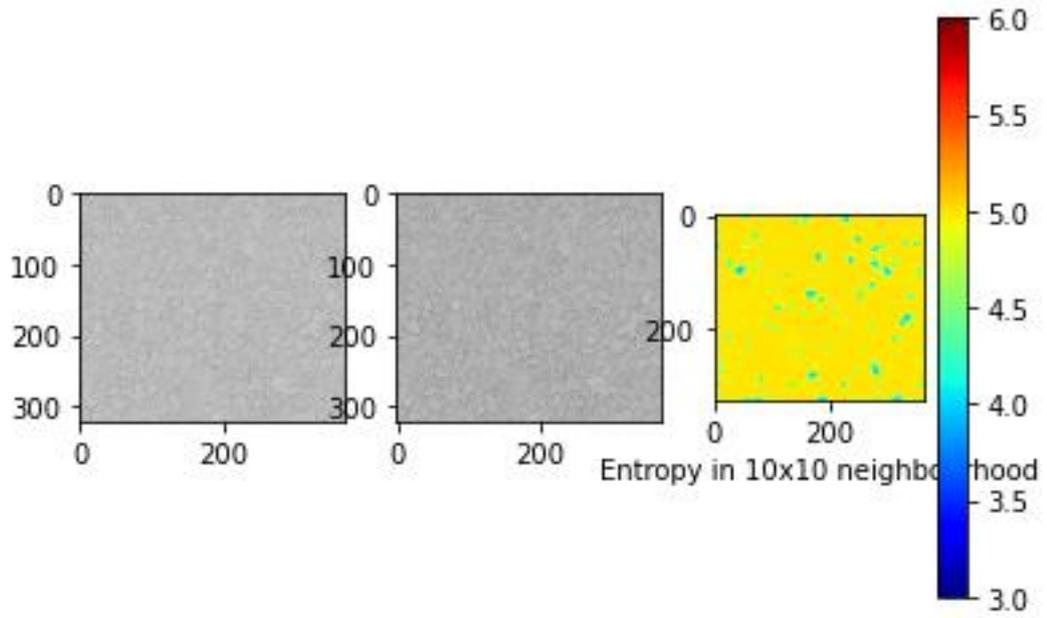

Figure 12: Entropy image of single-phase brass Friction Stir Welded joint

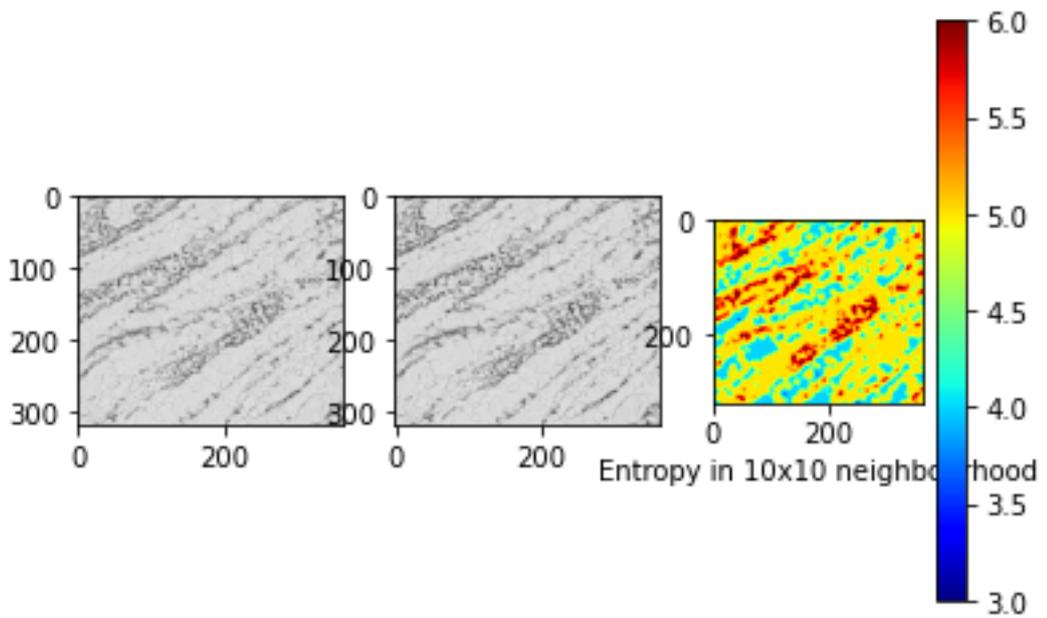

Figure 13: Entropy image of dual-phase brass Friction Stir Welded joint

RMS contrast stands for Root Mean Square contrast. For the recent research work, RMS contrast was preferred over other available contrast because it is a standard model for contrast measurement, and it is also applied in cortical cell responses for contrast normalization models. When the human eyes explore their surrounding then there is a variation in the local contrast and local luminance in the receptive field of a given visual neuron change.

Cosine weighting function as shown in equation 6 is used to measure the local contrast and luminance. In the given equation 'p' is the path radius, $(x_i, y_i)$ is the location of the $ith$ pixel in the given patch, and $(x_c, y_c)$ is the location center of the given patch.

$$w_i = 0.5 \left( \cos\left(\frac{\pi}{p}\sqrt{(x_i - x_c)^2 + (y_i - y_c)^2}\right) + 1 \right) \tag{6}$$

The local luminance of the given patch is obtained by the equation 7. In the given equation, $w_i$ is the raised cosine weighting function, N is the total number of pixels in the patch, and $L_i$ is the luminance of the $ith$ pixel.

$$L = \frac{1}{\sum_{i=1}^{N} w_i} \sum_{i=1}^{N} w_i L_i \tag{7}$$

The RMS contrast of the given patch is defined by the equation 8.

$$C_{rms} = \sqrt{\frac{1}{\sum_{i=1}^{N} w_i} \sum_{i=1}^{N} w_i \frac{(L_i - L)^2}{L^2}} \tag{8}$$

From the Table 1 it is inferred that the RMS contrast value of the dual-phase brass microstructure is high in comparison to other microstructures.

The Mean value of an image defines the contribution of individual pixel intensity for the entire image and it is calculated by the ratio of the sum of pixel values to the total number of pixel values. Equation 9 represents the method of calculating the mean value of the given microstructure image where $v_i$ is the sum of pixel value and $n$ is the total number of pixel values.

$$\bar{v} = \frac{1}{n} \sum_{i=1}^{n} v_i \tag{9}$$

From Table 1 it is observed that the dual-phase brass has higher mean value of 186.075 in comparison to other microstructures.

Table 1: Quantitative Metrics features of the reference microstructures

| Microstructures | Colorfulness Metric Value | Entropy | RMS Contrast | Mean |
|---|---|---|---|---|
| Single-Phase Brass | 0.118 | 7.504 | 46.000 | 183.885 |
| Dual-Phase Brass | 0.133 | 7.514 | 51.693 | 194.826 |
| Single-Phase Brass Friction Stir Welded Joint | 0.0 | 6.138 | 18.698 | 186.075 |
| Dual-Phase Brass Friction Stir Welded Joint | 0.0 | 6.437 | 31.978 | 202.998 |

The reference microstructures are converted to different color spaces i.e. XYZ color space as shown in Figure 14, and YUV color space as shown in Figure 15.

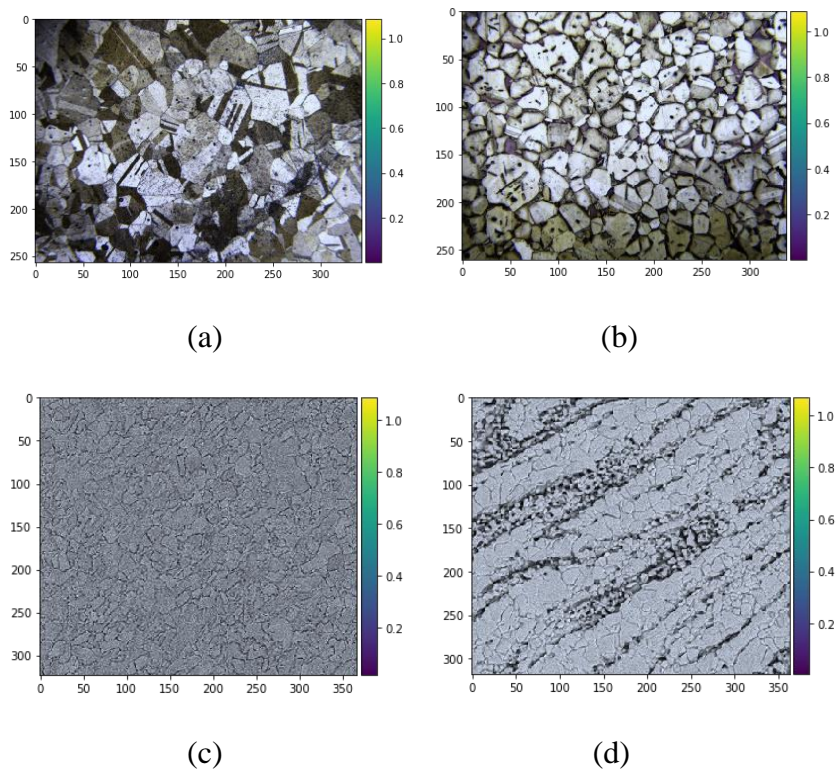

(a)  (b)

(c)  (d)

Figure 14: Obtained XYZ color space microstructure for a) Single-phase brass b) Dual-phase brass c) Single-phase brass Friction Stir Welded Joint d) Dual-Phase brass Friction Stir Welded joint

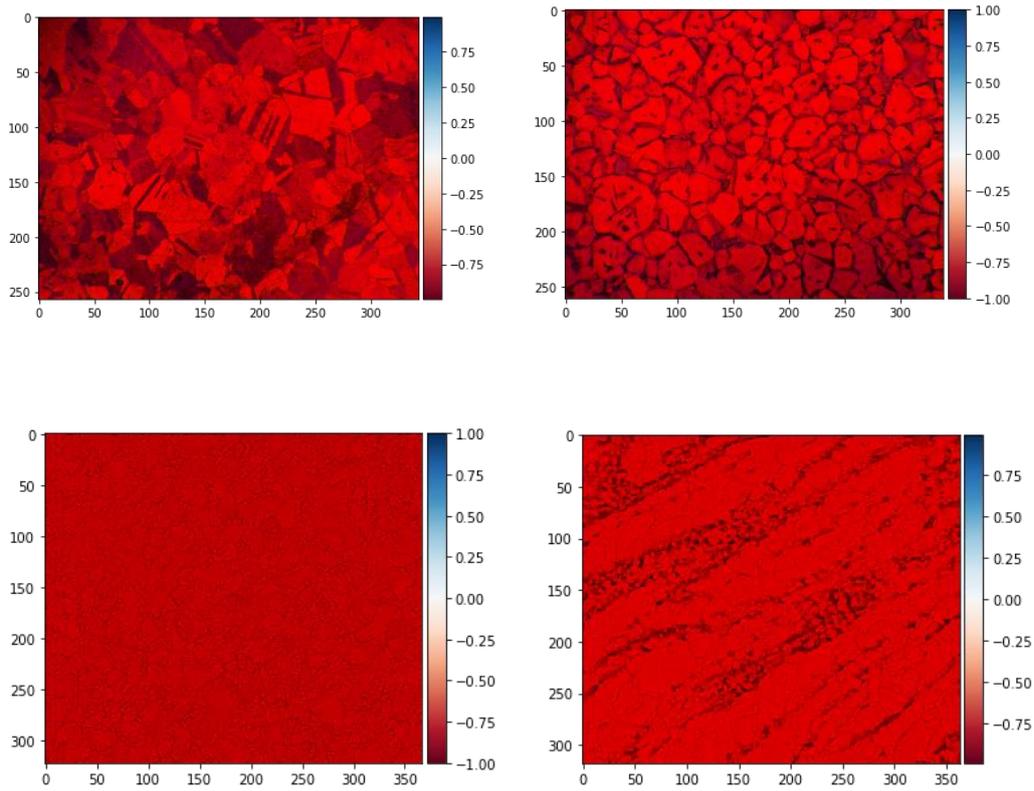

Figure 15: Obtained YUV color space microstructure for a) Single-phase brass b) Dual-phase brass c) Single-phase brass Friction Stir Welded Joint d) Dual-Phase brass Friction Stir Welded joint

Figure 16 to Figure 19 shows the Entropy of the reference microstructures in a XYZ color space. Table 2 shows the obtained quantitative metrics features of the reference microstructures in XYZ color space.

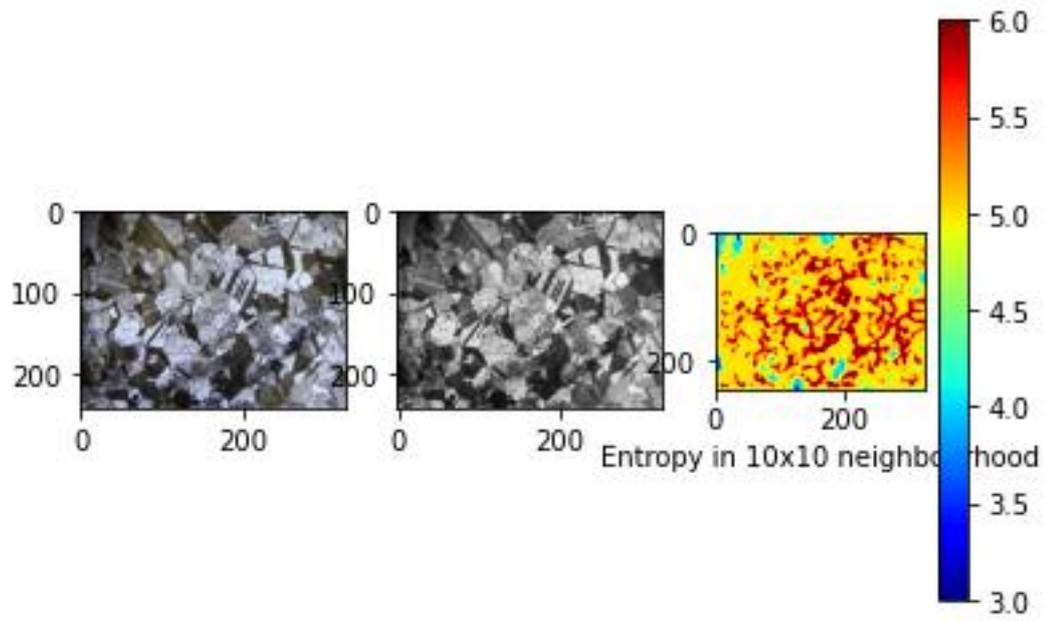

Figure 16: Obtained Entropy of the single-phase brass microstructure in XYZ color space

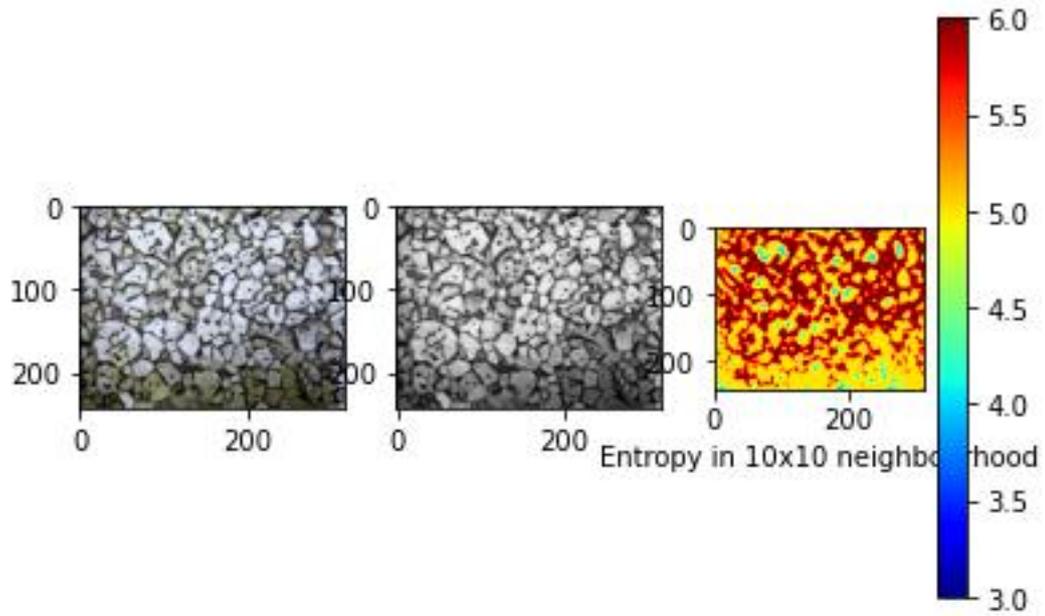

Figure 17: Obtained Entropy of the dual-phase brass microstructure in XYZ color space

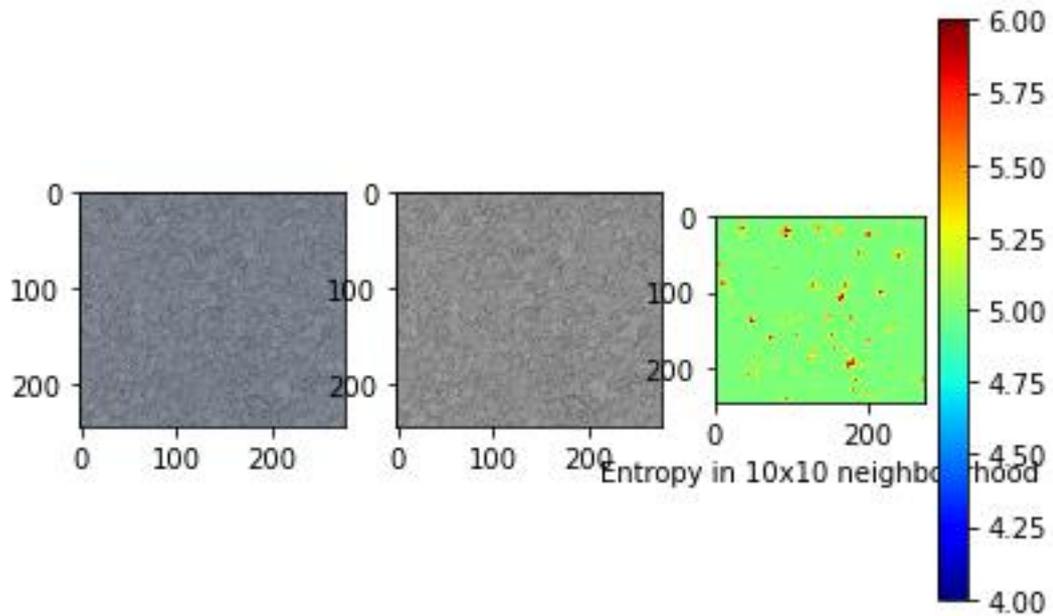

Figure 18: Obtained Entropy of the single-phase brass Friction Stir Welded microstructure in XYZ color space

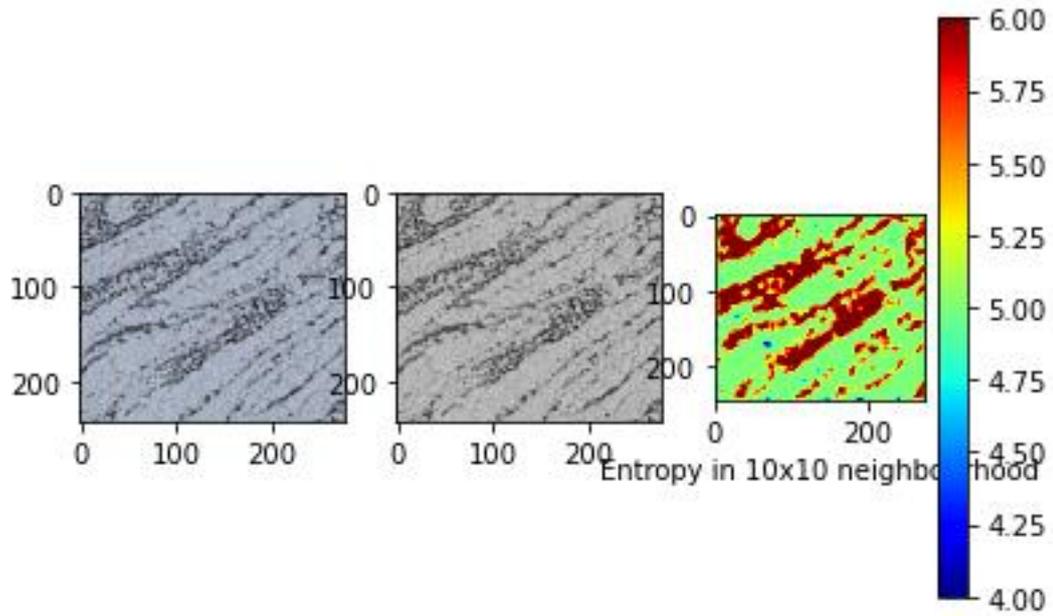

Figure 19: Obtained Entropy of the dual-phase brass Friction Stir Welded microstructure in XYZ color space

Table 2: Quantitative metrics features of reference microstructures in XYZ color space

| Microstructures | Colorfulness Metric Value | Entropy | RMS Contrast | Mean |
|---|---|---|---|---|
| Single-phase brass microstructure | 0.082 | 7.855 | 60.729 | 112.604 |
| Dual-phase brass microstructure | 0.085 | 7.955 | 67.184 | 129.905 |
| Single-phase brass Friction Stir Welded joint | 0.023 | 6.804 | 26.847 | 119.578 |
| Dual-phase brass Friction Stir Welded joint | 0.038 | 7.278 | 43.622 | 149.319 |

After comparing the results obtained in Table 1 and 2, it is observed that the Colorfulness Metric Value, Entropy , and RMS Contrast value in XYZ color space has higher value than the reference microstructures images shown in Table 1 while the mean value of the microstructures in XYZ color space is less than the mean values of the original reference microstructures.

Figure 20 to Figure 23 shows the entropy of the reference microstructures in YUV color space.

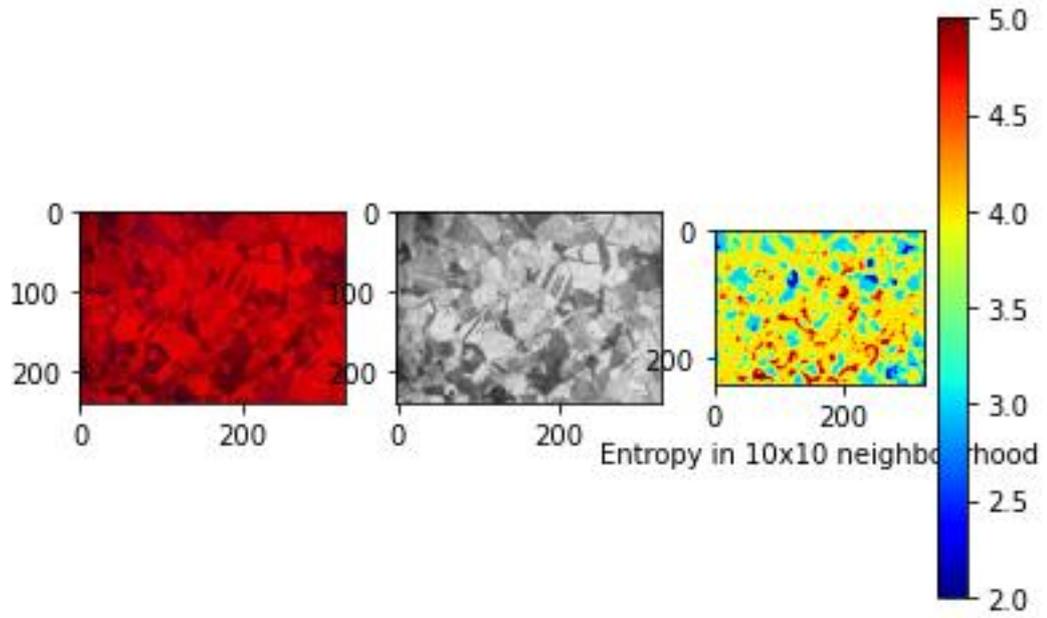

Figure 20: Obtained Entropy of the single-phase brass microstructure in YUV color space

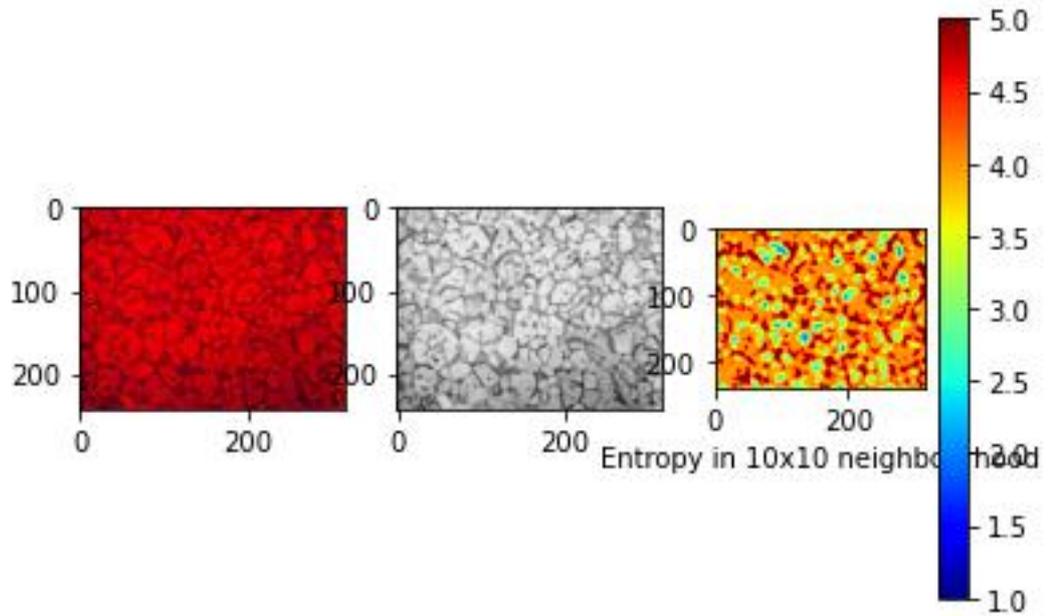

Figure 21: Obtained Entropy of the dual-phase brass microstructure in YUV color space

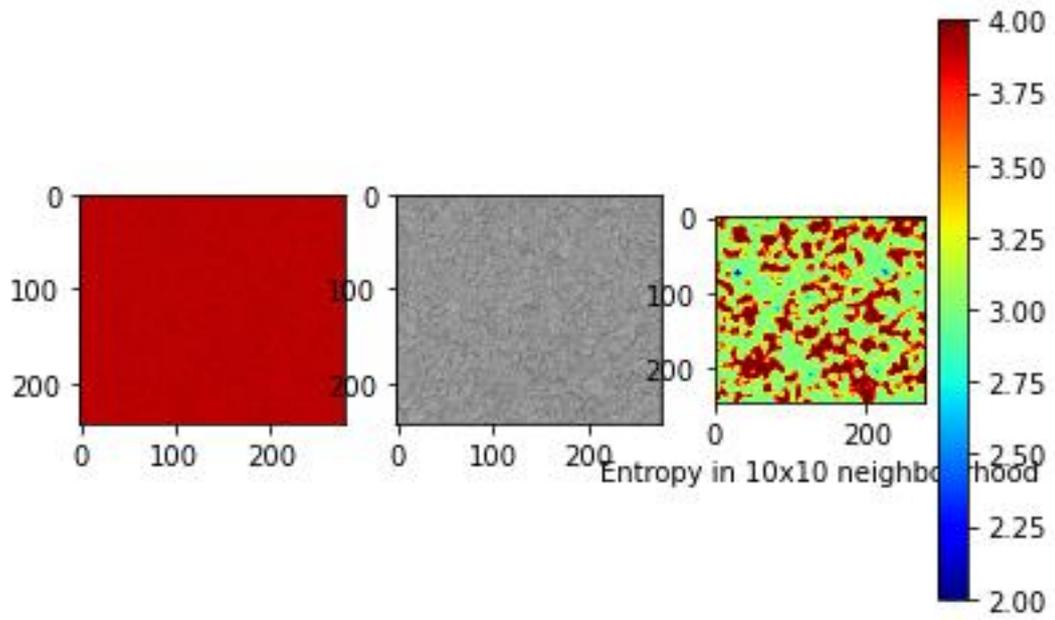

Figure 22: Obtained Entropy of the single-phase brass Friction Stir Welded microstructure in YUV color space

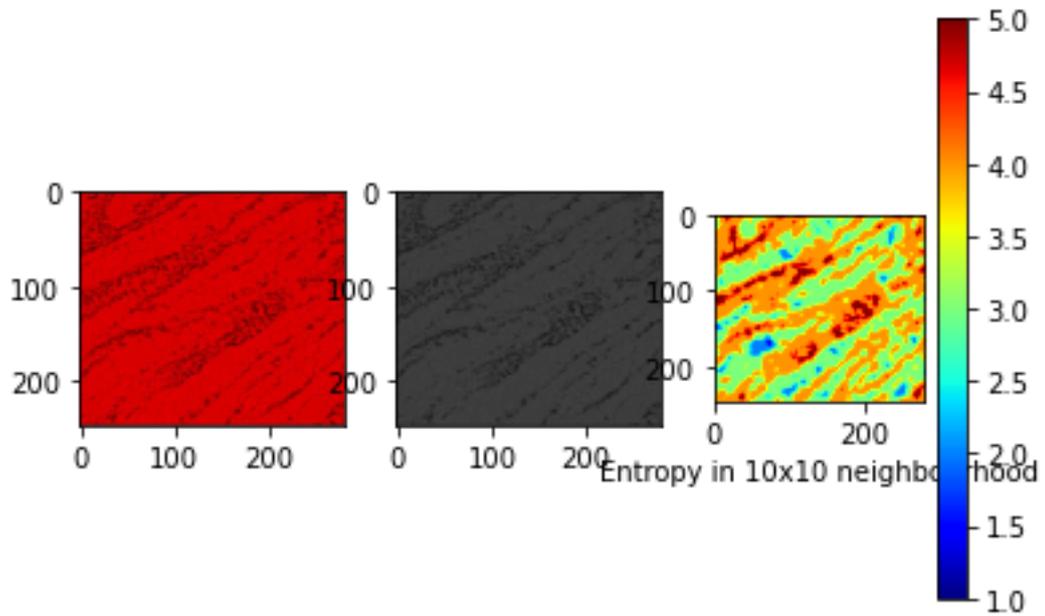

Figure 23: Obtained Entropy of the dual-phase brass Friction Stir Welded microstructure in YUV color space

Table 3: Quantitative metrics features of reference microstructures in YUV color space

| Microstructures | Colorfulness Metric Value | Entropy | RMS Contrast | Mean |
|---|---|---|---|---|
| Single-phase brass microstructure | 0.422 | 6.179 | 12.822 | 171.228 |
| Dual-phase brass microstructure | 0.459 | 6.203 | 13.530 | 181.899 |
| Single-phase brass Friction Stir Welded joint | 0.308 | 4.585 | 4.059 | 185.883 |
| Dual-phase brass Friction Stir Welded joint | 0.408 | 4.887 | 14.487 | 202.064 |

It is observed that the Colorfulness Metric in YUV color space is higher than the XYZ color space and the reference microstructures. It is also observed that the entropy, RMS contrast, and mean of the microstructures in YUV color space is less in comparison to the XYZ color space and the reference microstructures.

For enhancing the contrast in an image, the Histogram equalization technique which is an image processing technique is used. Following Python code is used for obtaining the histogram image of single-phase brass microstructure which is shown in Figure 24.

```python
import numpy as np

import cv2 as cv

from matplotlib import pyplot as plt

img = cv.imread('F:/CLAHE PROJECT/sb brass.jpg',0)

hist,bins = np.histogram(img.flatten(),256,[0,256])

cdf = hist.cumsum()

cdf_normalized = cdf * float(hist.max()) / cdf.max()

plt.plot(cdf_normalized, color = 'b')

plt.hist(img.flatten(),256,[0,256], color = 'r')

plt.xlim([0,256])

plt.legend(('cdf','histogram'), loc = 'upper left')

plt.show()
```

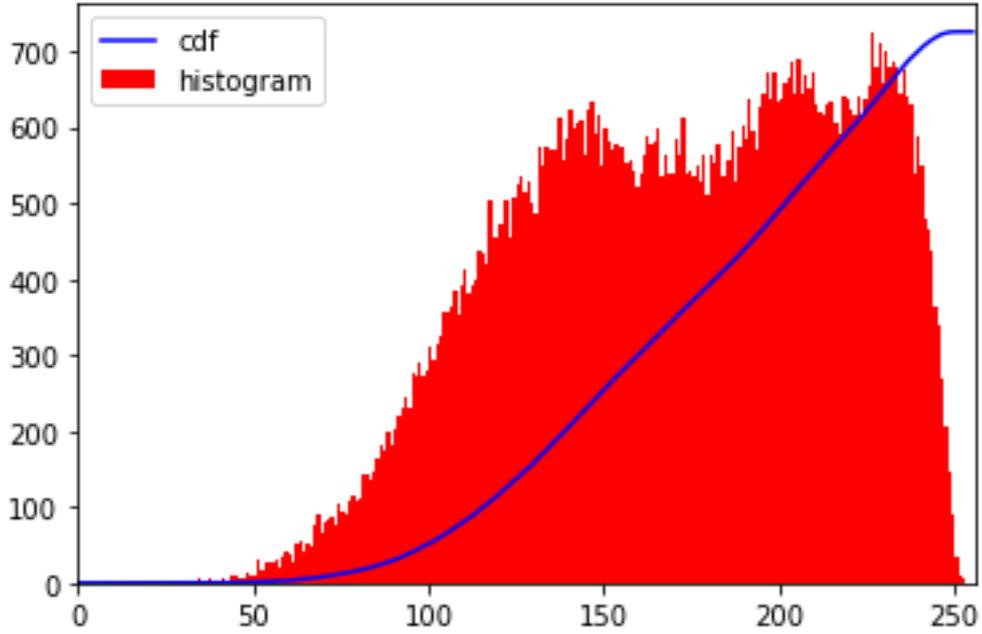

Figure 24: Histogram obtained for single-phase brass microstructure (Corresponding histogram (red) and cumulative histogram (blue))

It is observed from the Figure 24 that the histogram lies in the brighter region. The obtained cumulative histogram is governed by the equation 10 where, $p_x$ the probability of an occurrence of a pixel of level i in the image. So in order to obtain the full spectrum, histogram equalization is a transformation function used for mapping the input pixels in brighter regions to output pixels in the full region which is represented by the general histogram equalization shown in equation 11.

$$cdf_x(i) = \sum_{j=0}^{i} p_x(x = j) \qquad (10)$$

$$h(v) = round\left(\frac{cdf(v) - cdf_{min}}{(M \times N) - cdf_{min}} \times (L - 1)\right) \qquad (11)$$

where, $cdf_{min}$ is the minimum non-zero value of the cumulative distribution function, M × N gives the image's number of pixels and L is the number of grey levels used.

The Python code used for Histogram Equalization is shown below and obtained plot is shown in Figure 25.

```
cdf_m = np.ma.masked_equal(cdf,0)

cdf_m = (cdf_m - cdf_m.min())*255/(cdf_m.max()-cdf_m.min())

cdf = np.ma.filled(cdf_m,0).astype('uint8')

img2 = cdf[img]

hist,bins = np.histogram(img2.flatten(),256,[0,256])

cdf = hist.cumsum()

cdf_normalized = cdf * float(hist.max()) / cdf.max()

plt.plot(cdf_normalized, color = 'b')

plt.hist(img2.flatten(),256,[0,256], color = 'r')

plt.xlim([0,256])

plt.legend(('cdf','histogram'), loc = 'upper left')

plt.show()
```

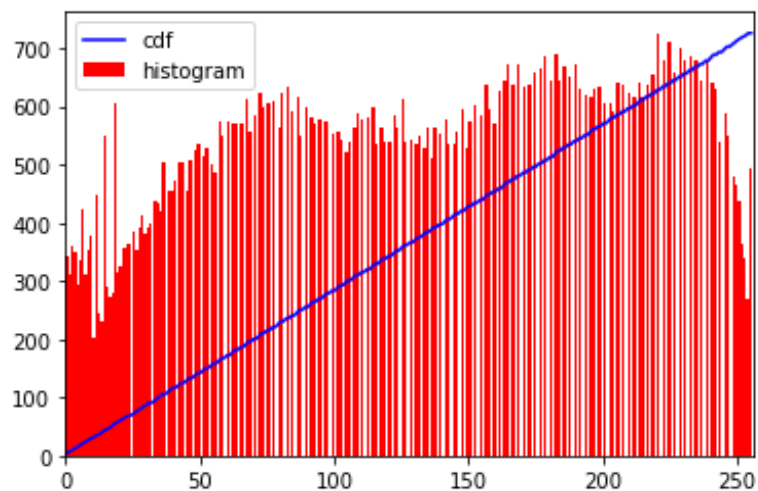

Figure 25: After Histogram equalization (Corresponding histogram (red) and cumulative histogram (blue))

The limitations of the Histogram equalization method are that it cannot be implemented in the images which have large intensity variation. So, in order to overcome this limitation Contrast Limited Adaptive Histogram Equalization (CLAHE) algorithm is used in the present work.

The Python code used for executing the CLAHE algorithm on original microstructures is shown below and the obtained enhanced microstructures are shown in Figure 26 and 27.

```
import numpy as np

import cv2 as cv

img = cv.imread('F:/CLAHE PROJECT/sb brass.jpg',0)

# create a CLAHE object (Arguments are optional).

clahe = cv.createCLAHE(clipLimit=2.0, tileGridSize=(8,8))

cl1 = clahe.apply(img)

cv.imwrite('F:/clahe1.jpg',cl1)
```

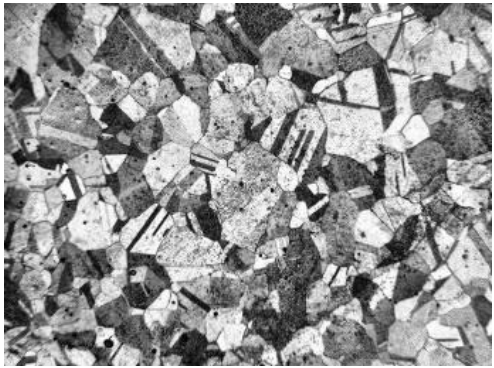 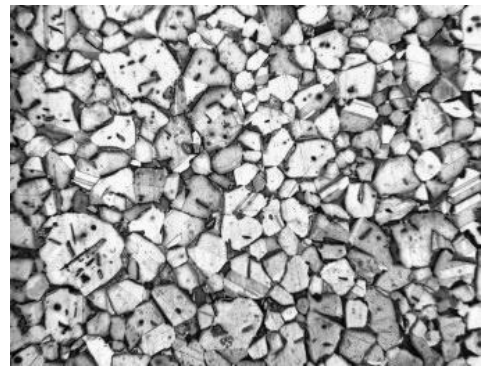

(a)  (b)

Figure 26: CLAHE enhanced microstructures of a) Single-phase brass microstructure b) dual-phase brass microstructure

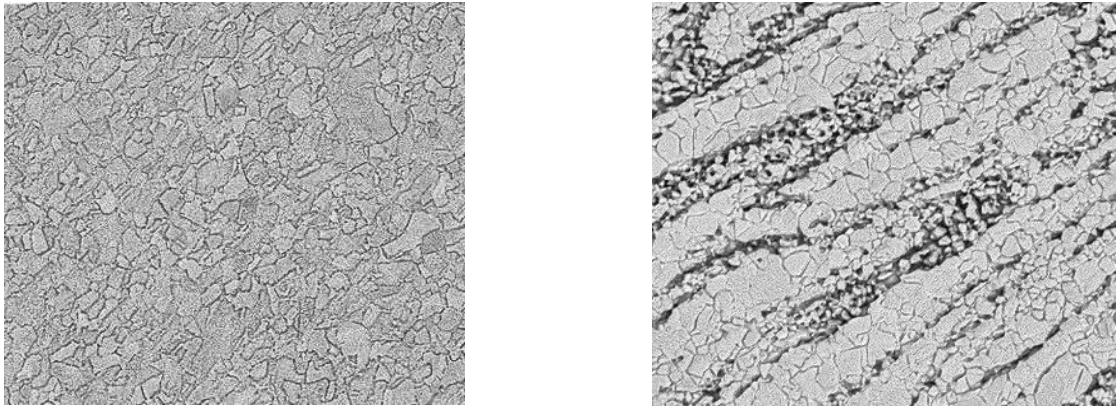

Figure 27: CLAHE enhanced microstructures of a) Single-phase Friction Stir Welded brass microstructure b) dual-phase friction stir welded brass microstructure

The obtained entropy features for given CLAHE enhanced microstructures are shown in Figure 28-31 and further obtained quantitative metric features are shown in Table 4.

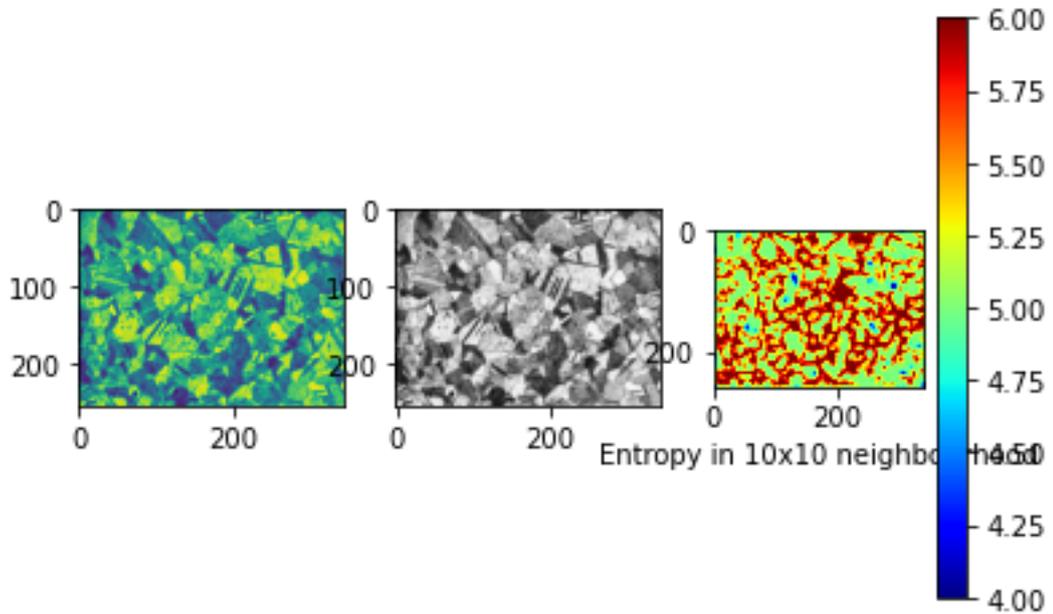

Figure 28: Obtained entropy image of CLAHE enhanced single-phase brass microstructure

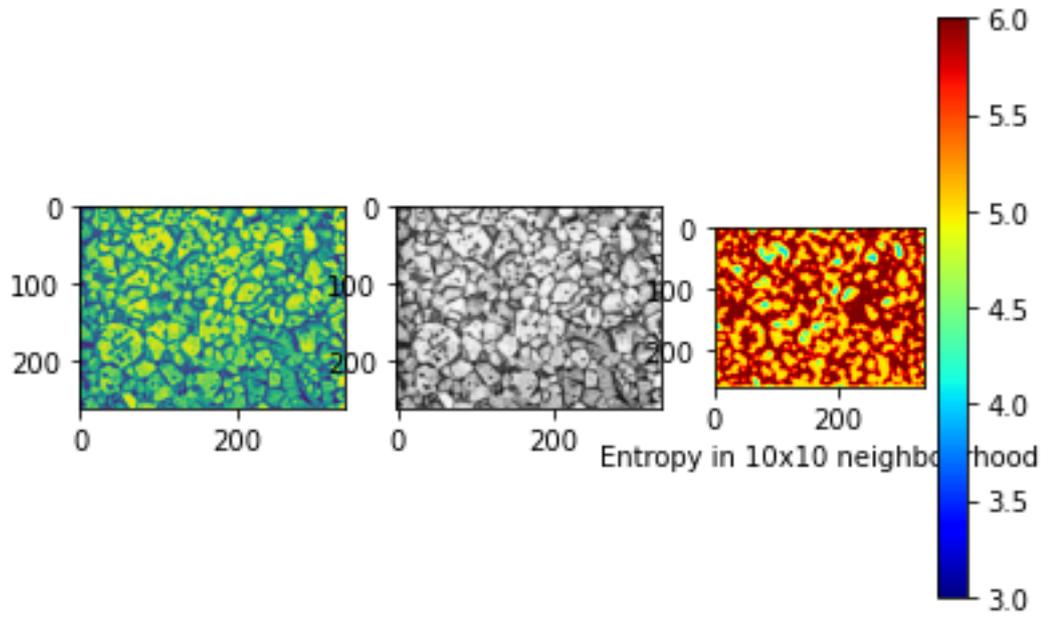

Figure 29: Obtained entropy image of CLAHE enhanced dual-phase brass microstructure

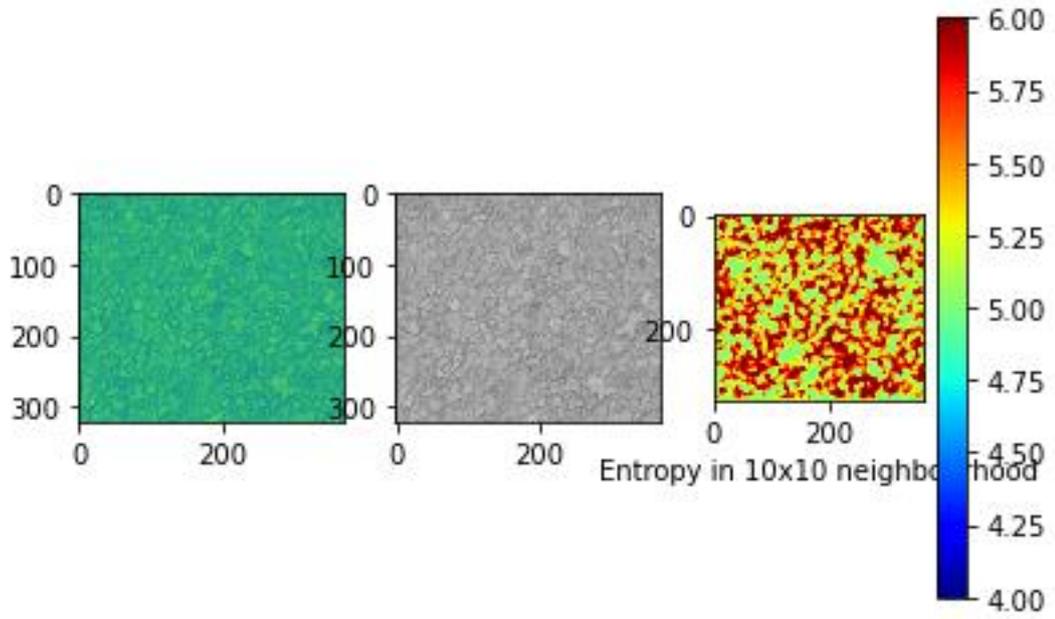

Figure 30: Obtained entropy image of CLAHE enhanced single-phase Friction Stir Welded brass microstructure

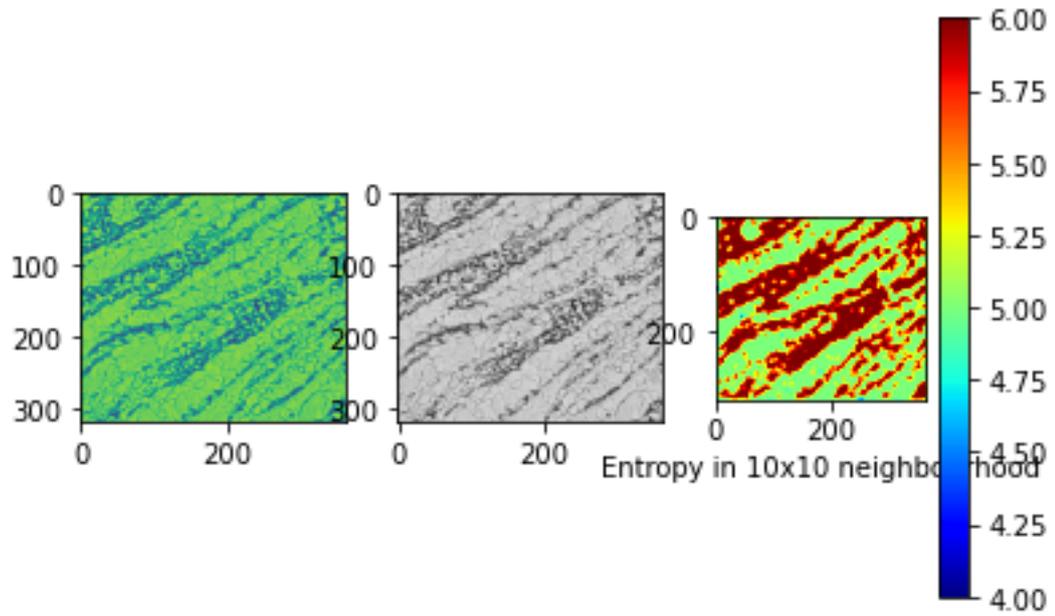

Figure 31: Obtained entropy image of CLAHE enhanced dual-phase Friction Stir Welded brass microstructure

Table 4: Quantitative metrics features of CLAHE algorithm enhanced microstructures

| Microstructures | Colorfulness Metric Value | Entropy | RMS Contrast | Mean |
| --- | --- | --- | --- | --- |
| **Single-phase brass microstructure** | 0.0 | 7.875 | 64.721 | 145.799 |
| **Dual-phase brass microstructure** | 0.0 | 7.832 | 67.765 | 158.083 |
| **Single-phase brass Friction Stir Welded joint** | 0.0 | 7.243 | 38.741 | 168.782 |
| **Dual-phase brass Friction Stir Welded joint** | 0.0 | 7.393 | 50.162 | 177.875 |

From Table 4, it is clearly observed that the CLAHE algorithm enhanced microstructures images result higher value of entropy and RMS Contrast value in comparison to other used algorithms.

# 6. Conclusion

In the present work, CLAHE algorithm is successfully implemented for enhancing the quality of the microstructure images used for the study purpose. It is clearly observed that the obtained values of the quantitative metric features i.e. Entropy value and RMS Contrast value were high in comparison to the other microstructures used in various color space. So, it can be concluded that high entropy and RMS contrast value of the enhanced CLAHE microstructures are of high quality in comparison to their parent microstructures.